\documentclass[10pt,twocolumn,letterpaper]{article}

\usepackage{cvpr}
\usepackage{times}
\usepackage{epsfig}
\usepackage{graphicx}
\usepackage{caption}
\usepackage{subcaption}
\usepackage{amsmath}
\usepackage{amssymb}
\usepackage{booktabs}
\usepackage{chngcntr}
\usepackage{multirow}
\usepackage{tabularx}
\usepackage{subcaption}
\usepackage{float}
\usepackage{algpseudocode,algorithm,algorithmicx}

\usepackage{authblk}

\algrenewcommand\algorithmicrequire{\textbf{Input:}}
\algrenewcommand\algorithmicensure{\textbf{Output:}}
\cvprfinalcopy % *** Uncomment this line for the final submission
\usepackage[breaklinks=true,bookmarks=false]{hyperref}

\def\cvprPaperID{****} % *** Enter the CVPR Paper ID here

\begin{document}

%%%%%%%%% TITLE
\title{Pixel-wise object tracking using a global attention model and a local segmentation model}

%\author{Yilin\\
%% omit the following lines up until the closing ``}''.
%% Additional authors and addresses can be added with ``\and'',
%% just like the second author.
%% To save space, use either the email address or home page, not both
%\and
\author[1]{Yilin Song }
\author[1]{Chenge Li }
\author[1]{Yao Wang }
\affil[1]{Department of Electrical and Computer Engineering,  New York University, NY, USA}

\maketitle
%\thispagestyle{empty}

%%%%%%%%% ABSTRACT
\begin{abstract}
In this paper, we propose a novel pixel-wise visual object tracking framework that can track any anonymous object in a noisy background. The framework consists of two submodels, a global attention model and a local segmentation model. The global model generates a region of interests (ROI) that the object may lie in the new frame based on the past object segmentation maps; while the local model segments the new image in the ROI. Each model uses a LSTM structure to model the temporal dynamics of the motion and appearance, respectively. To circumvent the dependency of the training data between the two models, we use an iterative update strategy. Once the models are trained, there is no need to refine them to track specific objects, making our method efficient compared to online learning approaches. We demonstrate our real time pixel-wise object tracking framework on a challenging VOT dataset.

% The appearance of the object and the location of the object are each considered as a time series and sepearately modeled. To close the observation gap between these two models, we use an iterative update strategy. Once the models are trained, there is no need to refine for specific objects, which makes our method efficient compare to online learning approaches. We further demonstrate our real time pixel object tracking framework on a challenging dataset. 

\end{abstract}

%%%%%%%%% BODY TEXT
\section{Introduction}
Provided with an object of interest at the first frame, visual object tracking is a problem of buiding a computational model that is able to predict the location of the object in consecutive frames. A robust tracking algorithm should be able to tackle some of the common issues including: target deformation, motion blur, illumination change, partial occlusion and background clutters. Many existing algorithms uses online learning by building a discriminative model to seperate object from background. The feature extractor is the most important component of a tracker, using appropriate features could drasticly boost the tracking performance. Many recent tracking-by-detection approaches \cite{pmlr-v37-hong15, wang2016stct,ning2017spatially} are inspired by methods for object detection \cite{girshick2015fast,redmon2016you,ren2015faster} and fully embrace the features learnt from deep convolutional neural network. We recognize that existing CNN based feature extractor increases the performance and robustness of the tracking system, yet how to extend the deep neural network for visual object tracking has not been fully investigated. In our work, we tackle object tracking as a time-series prediction problem, in particular we want to give a pixel-wise foreground-background label for consecutive frames.

Segmentation-based tracking algorithms \cite{aeschliman2010probabilistic,belagiannis2012segmentation, son2015tracking,yeo2017superpixel,jang2017online} have advantage over detection-based algorithm for handling a target undergoes substantial no-rigid motions. Many of them \cite{belagiannis2012segmentation, son2015tracking, aeschliman2010probabilistic} rely only on pixel-level information 
 and hence fail to consider semantic structure of the target. \cite{yeo2017superpixel} uses Markov Chain on superpixel graph, but information propagation through a graph could be slow depending on the structure. \cite{jang2017online} uses a encoder-decoder sturcture which shares some similarity with ours, but they rely on optical flow and markov random field, which limits the segmentation speed to around 1 fps and high image quality dataset as DAVIS\cite{perazzi2016benchmark}. The encoder-decoder structure is widely used in deep learning systems \cite{makhzani2015adversarial, kingma2013auto,pinheiro2015learning,yang2016object}. \cite{pinheiro2015learning,yang2016object} uses deconvolution for image segmentation and contour detection. In our work, we use a decoder to directly perform pixel-wise classification (object or not) on a video sequence.

To consider the target appearance variation in object tracking, several recent trackers embed CNN into their frameworks. Specifically \cite{tao2016siamese,bertinetto2016fully} modify siamese network structure for visual tracking purpose. \cite{wang2015transferring} trains a CNN using Imagenet data and transfers rich features learnt to a new object sequence by updating the network in an online manner. \cite{nam2016learning} trained a multi-domain network and has seperate branches for different domain sequences. The domain specific layer needs to be refined for each sequence. One major limitation of the aforementioned methods is that they lack mechanism to jointly model spatial-temporal traits of the object.  \cite{gan2015first,kahou2015ratm,yang2017recurrent} propose to solve tracking problem as sequential postion prediction by training RNN to model the time series. \cite{gan2015first,kahou2015ratm} uses RNN to model the temporal relationship among frames, but they only conducted experiments on synthesized data and did not demonstrate competitive result on challenging dataset like VOT \cite{unknown}. 
\cite{romera2016recurrent} uses convolutional LSTM \cite{xingjian2015convolutional} (convLSTM) to perform instance segmentation on single image. By spatial inhibition with an attention mechanism, they demonstrate compelling result on VOC Pascal dataset \cite{everingham2015pascal}. \cite{yang2017recurrent} uses convLSTM to model object feature variations, and their object detection mechanism use similar convolution structure as \cite{bertinetto2016fully}. By performing convolutional operation between exemplar frame with a region of interest (ROI), the output of their system is too coarse for fine-grained pixel labeling.

%Our pixel object tracking network is semi-supervised model. Provided with the ground truth segmentation for the object at the first frame, our system generates object segmentation results in real-time for the following frames.The segmenation is done on a ROI sampled from the raw frame. We use a temporal attention mechanism to locate the ROI. We tested our framework on a very challenging dataset VOT 2016\cite{unknown} to evaluate our pixel-wise tracking result. In summary, our main contribution are three folds: 
%(i) 
We propose a novel object tracking framework consisting of two models. The global model learns the global motion pattern of the object and predicts the object's likely location in a new frame from its past locations. The local model performs object segmentation in a ROI identified by the global model, based primarily on the appearance features of the object in the new frame. The local model uses a convLSTM based structure whose memory state evolves to learn the essential appearance features of the object, enabling the segmentation of the object even under significant appearance shifts and occlusion. The LSTM output further goes through a deconvolution layer to generate the segmentation map. The global model also employs  a convLSTM structure to generate the latent feature characterizing the object motion, which is fed to a spatial transformer network to determine the location and size of the ROI in the new frame. The proposed framework has demonstrated promising performance on a very challenging dataset (VOT 2016 \cite{unknown}), where some objects are very small relative to the image sizes, and testing videos often contain unseen objects in the training videos (in our cross validation study).

\section{Framework}
\begin{figure}
        \centering
	
		\includegraphics[width=0.50\textwidth]{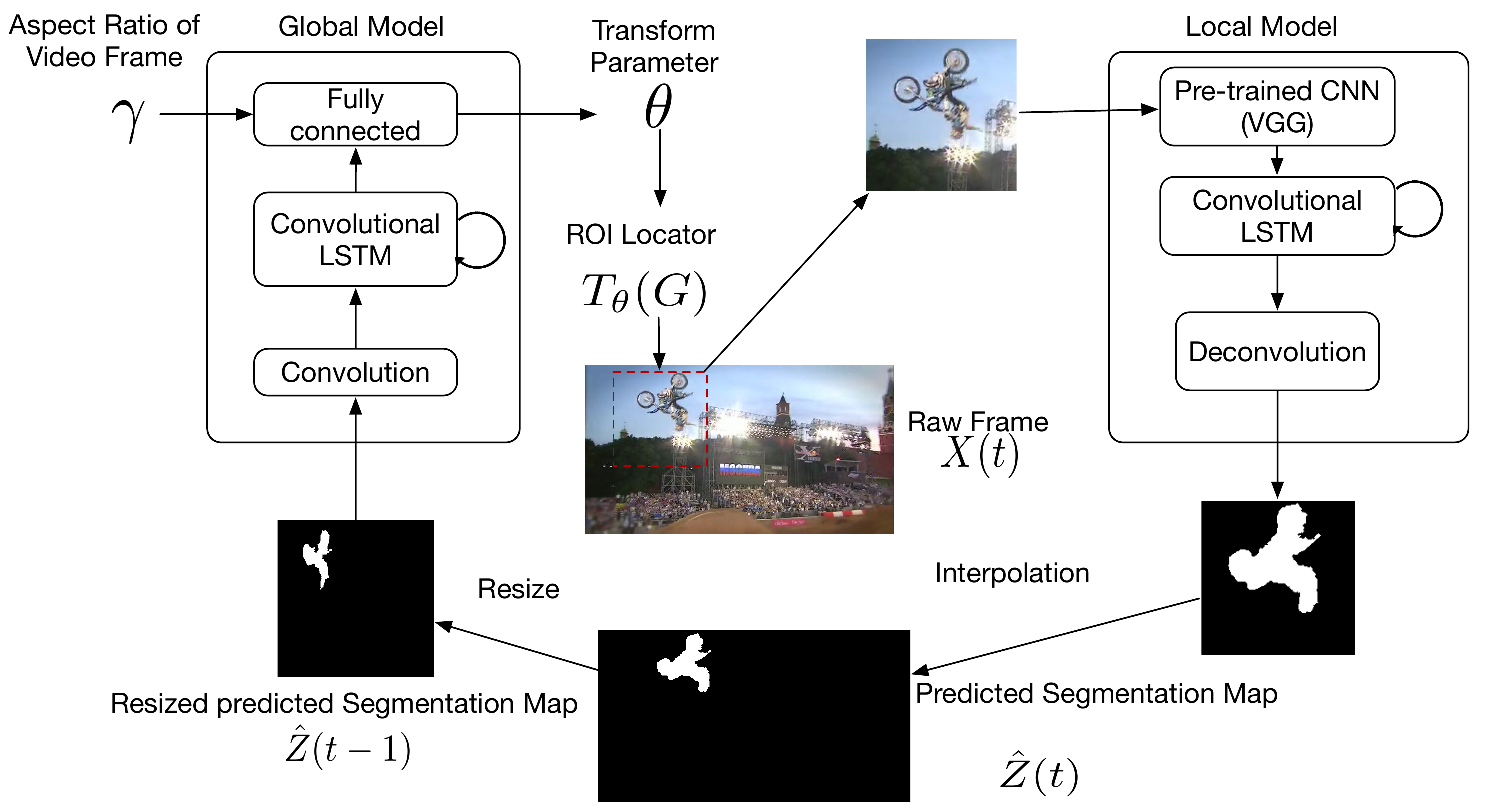}
		\caption{Pixel-wise object tracking framework. The network consists of two sub-modules: a local and a global model, working in a closed loop. At time step $t$, a resized full resolution binary image $\hat{Z}(t-1)$ is feed into the global model. In inference time, this binary image is the predicted segmentation map acquired from the local model at frame $t-1$. The global model then roughly predict where the object would appear in  frame $t$ based on past segmentation maps and generate a region of interests (ROI). The cropped image in the ROI at frame $t$ is then fed into the local model for segmentation.}

\label{Tracking framework}
\end{figure} 

 %This could be viewed as a special case of spatial transform network\cite{jaderberg2015spatial}, where the transform parameter is learnt through a recurrent network. The localization network learns transform parameter $\theta$ from series of resized square binary segmenation images. And grid sampler $T_\theta(G)$ then sample a image template from full resolution image with aspect ratio $\gamma$.

%
%
%As the grid sampler $T_\theta(G)$
%Another difference with spacial transform network is that the localization parameter $\theta$ and grid sample $T_\theta(G)$ is operating on different domain
%
%To accommodate for different image aspect ratio,   
%the last predicted segmenation map $\hat{Z}_{t-1} is feed into a global attention model. 
     
Our goal is to build a pixel-wise object tracking framework for all possible image resolutions and aspect ratios.  Models will be trained in an offline supervised setting. Once the offline training process is finished, there is no need to retrain the network. Segmentation of the full-resolution image would require a large amount of computation resources for real time application. In addition, scaling original image to a fixed size could  destroy the semantic information and appearance features of a small object relative to the image size. To overcome these difficulties, a global model is used to predict the rough location of the object based on the past object segmentation maps. We then crop a region of interest (ROI) from the original image and perform segmentation on the ROI. The network structure is shown in Fig.~\ref{Tracking framework}.

The model runs in a close loop during inference time.  At time step $t$ the global model takes a fixed size segmentation map as input, which is the resized version of a  predicted full resolution segmentation map derived at $t-1$. We use several layers of convolution and pooling to reduce the dimensionality of the image. The resulting features are fed into a convolutional LSTM \cite{xingjian2015convolutional} to fully exploit the temporal variation characteristics of the past segmentation maps. To allow different ROI sizes for the local model (necessary to handle different object sizes and  object size variation due to motion in the depth direction), another fully connected layer takes convolution LSTM output as its input to estimate the spatial transformation parameter $\theta$ (including translation and scaling) for the ROI locator, which applies the transformation on a reference anchor box $G$ to generate the ROI in the raw frame $X_t$. As the input segmentation map to global model is resized to $224\times 224$, we inject the aspect ratio of the original image $\gamma$ into the last fully connceted layer of the global model for generalization power among video sequences with different aspect ratios.

%To preserve the aspect ratio of the object in the ROI, we want the scaling in the horizontal and vertical directions to be the same and define the parameter $\theta$ as:
%\[
%\theta = \begin{bmatrix}
%    s       &0 & t_x \\
%    0 & s &t_y \\
%\end{bmatrix}
%\]

At time step $t$ the local model receives a ROI image $x_t$ croped from the full resolution image $X_t$. A pretrained VGG is used to extract features from the ROI image. These features are then fed into a convLSTM to model for appearance shift. Then the output of convLSTM goes through a deconvolution layer to generate the local segmentation map (which is a gray scale image, with the value at each pixel proportional to the estimated likelihood that the pixel belongs to the tracked object). Based on how the ROI is cropped from the full resolution image, the full resolution segmentation map is interpolated accordingly from the ROI segmentation map. Here we assume the ROI encloses the entire object hence all pixels outside the ROI are set to zero.

Global model and local model are trained alternatingly in an end-to-end supervised manner. Once the offline training process is finished, there is no need to online finetune the network based on the appearance of the target object, as in some prior work \cite{nam2016modeling,nam2016learning,wang2015transferring}. 
\begin{figure}
        \centering
	
		\includegraphics[width=0.5\textwidth]{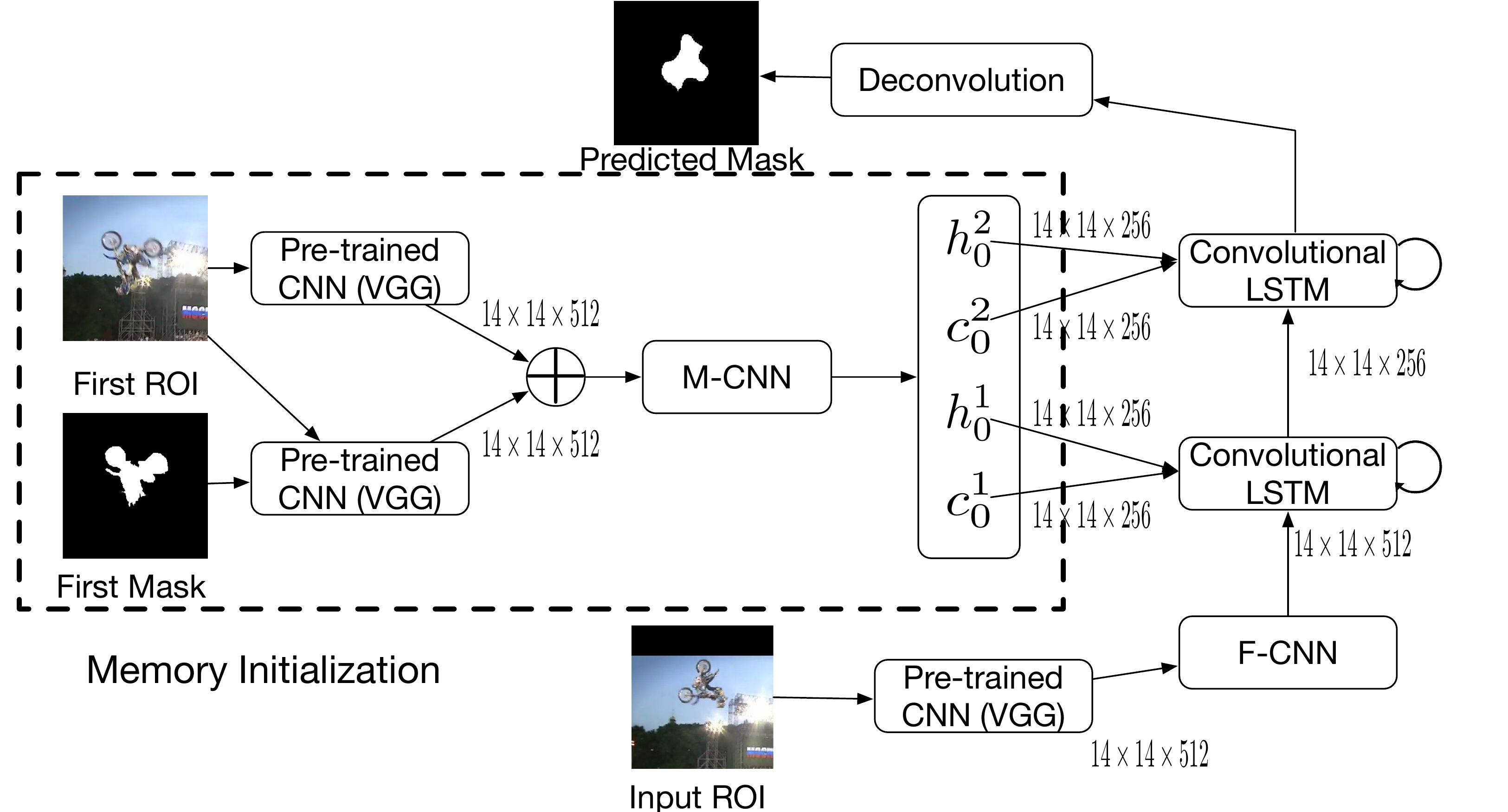}
		\caption{Local model for object segmentation in a ROI image. The M-CNN and F-CNN are feature normalization layers. $h,c$ are hidden and memory states. }

\label{Local framework}
\end{figure} 

\section{Local Segmentation Network} \label{Local segmenation network}

\subsection{Framework} \label{local_framework}
The network structure for our local model is shown in Fig.~\ref{Local framework}. We consider the pixel-wise object tracking as a time series prediction problem. At each time $t$, an input ROI $x_t$ is first processed by a pretrained convolutional network. As \cite{DBLP:journals/corr/ZeilerF13} has shown, high layer features of a trained CNN bears more semantic information whereas low layer outputs bears more appearance information. For genetic visual tracking, the features should be robust enough to work with many different object categories, and also be able to discriminate object instances from the same object class. Lower level features could be more helpful for such a task. However, using very low level features from a pretrained network could drasticly increase the computation cost for the following layers. Based on these consideration, we use pool4 features from a VGG network \cite{simonyan2014very} pretrained for image segmentation dataset. The weights of this feature extractor are kept the same during training.

The features are then fed into convolutional LSTMs. However, we have found that the resulting network is hard to train because pool4 features are not confined in a certain range. Therefore, we use another small network consisting of two convolutional layers to normalize the VGG features, which use $tanh$ as the last activation function. These parts are denoted as F-CNN in Fig.~\ref{Local framework}. The normalized features then go through a two layer convLSTM. Intuitively the first convLSTM layer models the dynamics of the foreground object as well as the background. And the second convLSTM layer mostly address appearance shift of the target object. The equation we use for ConvLSTM are shown in Eq.~(\ref{ConvLSTM}). To get the segmentation map, the output of second layer convolutional LSTM features are then fed into a deconvolution layer. 
\begin{equation}
\begin{aligned}
i_t &= \sigma ( W_{xi} \ast {X}_t + W_{hi} \ast H_{t-1} + W_{ci} \circ C_{t-1} + b_i) \\
f_t &= \sigma ( W_{xf}  \ast {X}_t + W_{fi}  \ast H_{t-1} +W_{cf} \circ C_{t-1}+b_f) \\
C_t &= f_t\circ C_{t-1}+i_t \circ tanh(W_{xc}\ast X_t+ W_{hc}\ast H_{t-1}+b_c) \\
o_t &= \sigma(W_{xo}  \ast {X}_t + W_{ho}  \ast H_{t-1} +W_{co }\circ C_{t}+b_o) \\
H_t &= o_t \circ tanh(C_t) \\
\end{aligned}
\label{ConvLSTM}
\end{equation}

In equation~\ref{ConvLSTM}, the hadamard product $\circ$ between $W_{c*}$ and $C$ are crucial for learning long term dependencies. It restricts cross-channel information exchange and overcomes vanishing gradient problem. Replacing hadamard product with convolution would not achieve similar performance for time sequence model. On the other hand, ConvLSTM is not equivariant to translation particluarly because of the hadamard product. This means a spatially shifted version of the input image may not lead to an equally shifted segmentation map. As the global model may not always generate the ROIs centered around the object at different frame times, it would be preferred that the local segmentation network has a certain degree of translation eqivariance. Although this is one major drawback of using ConvLSTM for object tracking, we have found that with the ROI chosen by the global model, the object tends to fall near the centers of the ROIs in all frames, and our local model can perform well even with small spatial shift between consecutive frames for unseen objects. The detailed number of parameters are shown in Tab.~\ref{num_parameters}.
\begin{table}
\scalebox{0.8}{
\begin{tabular}{  p{3cm} | p{2cm}  p{2cm} p{2cm} }
\toprule       
Local model &filter size &channels & stride \\
\toprule
M-CNN $\times 2$& $3\times3$  & 1024 &1 \\
F -CNN $\times 2$ & $3\times3$ & 512 & 1 \\
ConvLSTM $\times 2$ & $3\times3$ & 256 &1 \\
Decov -1 & $5\times5$ &128 & 2 \\
Decov -2 & $5\times5$ &64 & 2 \\
Decov -3& $5\times5$ &32 & 2 \\
Decov -4& $5\times5$ &1 & 2 \\
\toprule
Global model &filter size &channels & stride \\
\toprule
layer 1 $\times2$ & $3\times3$ &8& 1 \\
layer 2 $\times2$ & $3\times3$ &16& 1 \\
layer 3 $\times2$ & $3\times3$ &32& 1 \\
layer 4 $\times2$ & $3\times3$ &64& 1 \\
ConvLSTM$\times2$ &$3\times3$ &64 &1 \\
full 1 &  & 1024 & \\
full 2 &  & 3 & \\
\bottomrule

 \end{tabular}}
\caption{Number of filters for each modules in local segmentation network and global attention network. Notation $\times 2$ represents two identical layers that are connected. \textbf{Local model:} In M-CNN and F-CNN internal activation functions use rectified linear unit (relu), whereas the outut activation function is $tanh$. The internal activation function in deconvolution is leaky-relu, the last activation function is sigmoid. \textbf{Global model}: every two convolution layer are followed by a pooling operation to reduce the spatial dimentionality. The input to fully connected layer is vectorized ouput of convolutional LSTM. For the second fully connected layer the input dimension is 1025, where we concatenate the feature from last layer with aspect ratio of the current video clip.}
\label{num_parameters}
 \end{table}

\subsection{Memory Initialization}\label{Memory Initialization}
To start ConvLSTM, we need to initialize the memory and hidden state. Initializing the memory cell to be zeros is one option. But a major drawback of such  approach is the memory cell of recurrent network would need multiple time steps to converge. During this time its hidden connection $h$ is also drastically different from its true distribution.  And segmentation could easily fail because deconvolution is directly applied on $h$. A wrongfully predicted local segmentation map would further affect the global model. Moreover, within the first ROI there could be more than one salient object. Without differentiating between these salient objects, the tracking system would not know which object to track and is likely to fail.

\begin{figure}
        \centering
	\begin{subfigure}[b]{0.5\textwidth}
		\centering
		\includegraphics[width=1.0\textwidth]{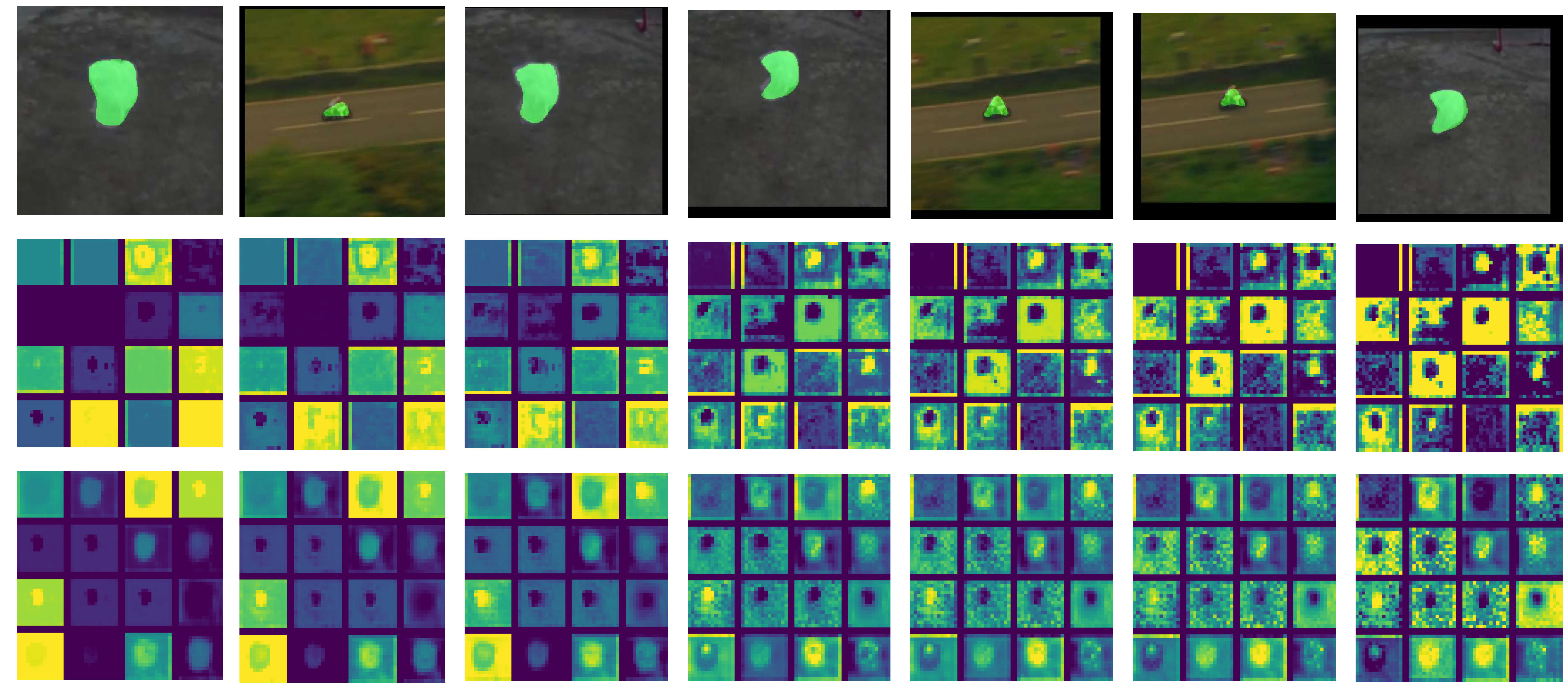}
		\caption{Sequential segmentation result visualization without randomly inserting noisy frames during training. }
	\end{subfigure} %
	\begin{subfigure}[b]{0.5\textwidth}
		\centering
		\includegraphics[width=1.0\textwidth]{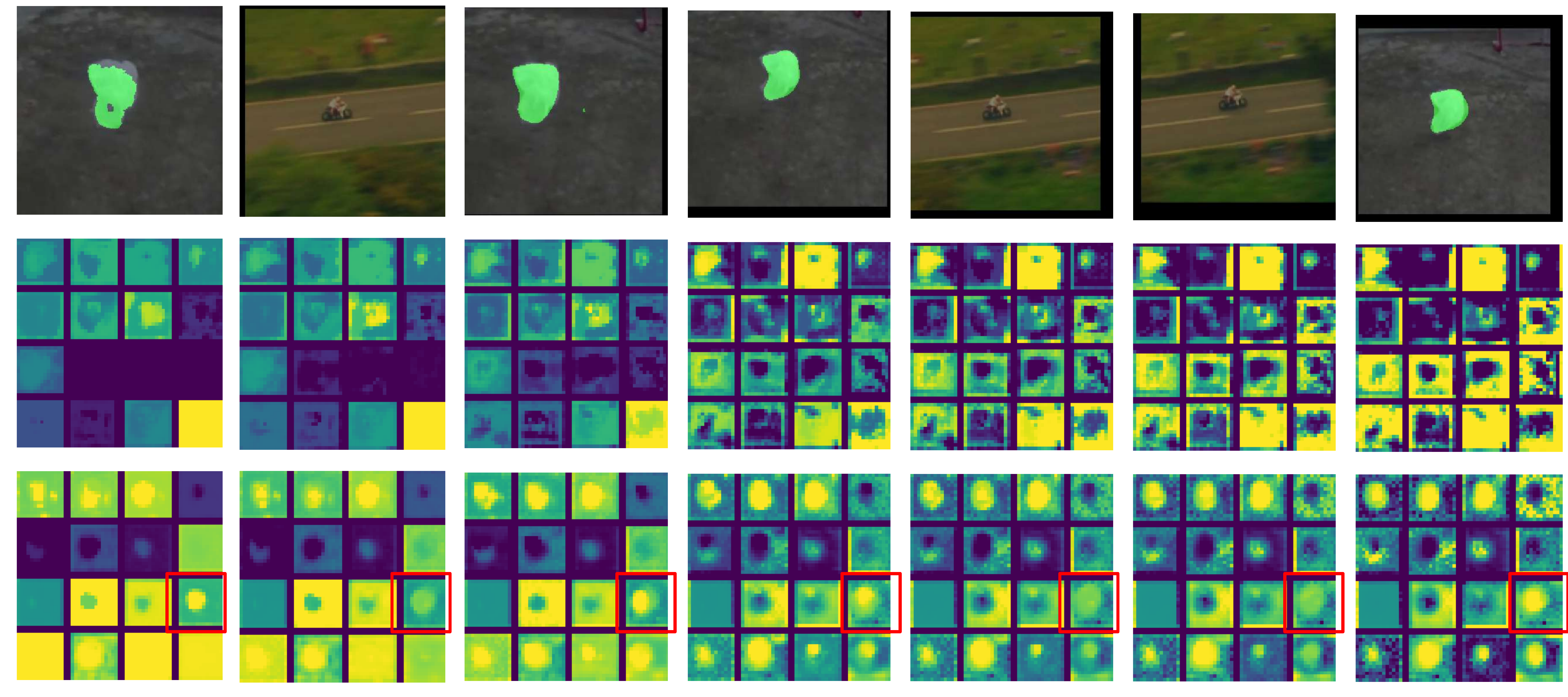}
		\caption{Sequential segmentation result visualization with randomly inserting noisy frames during training.}
	\end{subfigure}%

\caption{In each subfigure, each row in vertical order is: the segmentation result overlaid on top of the raw image, first layer convolutional LSTM memory cells, second layer convolutional LSTM memory cells. The displayed images are downsampled by 2. Row 1: Both true sequence and inserted frames comes from testing set. Row 2 and 3, we show the top 16 activations out of 256 cells. Note: (i) For ConvLSTM even with memory regression there is still a burn-in time for the memory to converge. (ii) Memory cells get far noiser in subfigure(a) compare to subfigure(b) after several steps. (iii) There is memory cells co-adapt with noisy sequences, which act as action detection (encircled with red rectangle in subfigure (b)). }
\label{memory_cell}
\end{figure}

Instead of arbitrarily initializing the memory with zero, we train an initialization module that takes the object mask, and the image in a manually chosen ROI in the first frame and generates the initial memory cell state and the hidden state which ideally should capture the appearance features of the object. To overcome the boundary artifact, we use a dilated mask to generate the masked image. In our experiment, we find that instead of applying the object mask in the image domain, applying the mask on the layer right before the pool1 layer in the VGG network would render better performance. We then regress the initial memory and hidden states of ConvLSTM using the concatenated feature. This is done by using another two convolution layers denoted by M-CNN in Fig.~\ref{Local framework}. Simiar as \cite{yang2017recurrent}, we find using a $tanh$ function as the last activation function for M-CNN stabilizes the memory, even thougth the numerical value of memory cell could go beyond the range of $[-1,1]$. Ideally we want the memory cell to slowly adapt to appearance drift meanwhile while being able to ignore false objects. In Fig.~\ref{memory_cell} we show the memory state evolution under different training strategies. The training strategies would be discussed in the following subsection.

\subsection{Training} \label{local_training}
Visual object tracking (VOT) \cite{unknown} dataset is considered one of the hardest dataset for object tracking, because it contains videos in varying resolutions and some of the target objects (e.g. a football) are very small relative to the image size, and some objects undergo significant appearance shifts. The dataset contains 60 video sequences with more than 200 frames per sequence on average. To deal with the limited number of videos, we use 10 fold-cross validation and randomly distribute 60 sequences into 10 data fold. Each fold contains 54 videos in the training set and the other 6 videos in the testing set. Testing videos often contains objects not seen in the training set. For all models training is only done on the training set and we report the average accuracy on the testing set. 

The minibatch of sequences are prepared by the following steps:

\begin{enumerate}
\setcounter{enumi}{0}
\item Manually select a frame from a sequence randomly as the initial frame. Initial frame does not contain artifacts including occlusion, motion blur etc.
%We randomly select a sequence. In each sequence, we randomly select a good frame where there is no occlusion, motion blur, lighting condition etc. as the initial frame. The "good" frames are manually labelled. 

\item Crop this and all subsequent frames to generate ground truth ROI images.  The width of the square ROI is twice the longer length of  the object along the horizontal and vertical directions. The ROI width is further truncated to within the range of $[56,672]$. In order to train the model to deal with the potential error of the global model, the location is set according to the  object mask at frame $t-5$. Resize all ROI images to $224 \times 224$, equal to the input image size for the VGG network.

\item Perturb the resulting ROIs in both positions and size randomly. Random scaling is set in the range of $[0.9,1.1]$ and spatial shift $[-10,10]$ pixels. We denote the resulting sequences of ROI images for all training videos (each video contains only one object) as $x_{0:T}$, and the sequences of ground truth segmentation masks within the ROI as $z_{0:T}$. 

\item For each training video $i$, replace the ROI image at a randomly chosen time $t_i$ with the ROI image for another randomly chosen video $j$ at another time $t_j$.  The ground truth segmentation maps for such ROI images are set to all zero. Motivation for this step is explained in Sec.~\ref{Comparison and analysis}.

\end{enumerate}

After these steps, each training sample is a pair of video clips (the ROI image sequence and the ground truth ROI mask sequence for a training video), we then solve the following optimization problem in Eq.~(\ref{optimization1}), where $L(\hat{z}_{1:t},z_{1:t})$ and $V(\hat{z}_{i})$ are element-wise cross entropy loss and image total variation loss respectively. $\Phi$ defines the local segmentation network and $\theta$ is the parameters belonging to $\Phi$. We use the image total variation loss $V$ to discourage the resulting segmentation map to contain multiple small isolated components. We intentionally avoid applying more complicated post-processing on the segmentation map using approaches like markov random field (MRF) to both reduce the computation complexity at the inference time and to enable end-to-end training. $\beta$ is a thresholding term that stablizes the training procedure especially at the beginning stage.  $\beta=1000$ and $\lambda = 1e-4$ was found to achieve the best performance.
\begin{equation}
\begin{aligned}
&\min\limits_{\theta} L(\hat{z}_{1:T},z_{1:T}) + \lambda \min(\beta, \frac{1}{T} \sum_{i=i}^T V(\hat{z}_{i})) \\
&\hat{z}_{1:t} = \Phi_\theta(z_0,x_{0:t}) \\
&L(p,y)= \sum \limits_i -(1-y_i)log(1-p_i)-y_i log(p_i) \\
&V(y) = \sum_{i,j} |y_{i+1,j} - y_{i,j}|+|y_{i,j+1}-y_{i,j}|
\label{optimization1}
\end{aligned}
\end{equation}
\subsection{Comparison and Analysis} \label{Comparison and analysis}
We found step 4 in the data preparation is crucial for the success of the local segmentation network. Without step 4, the convolution LSTM merely learns a frame by frame saliency detection. In Fig.~\ref{memory_cell}, we compare the memory state evolution for two networks with and without step 4 on unseen sequences. The module learnt with step 4 is much more stable especially when there are multiple salient objects in the same ROI.

\begin{figure}
%\begin{subfigure}[t]{.45\textwidth}
    \centering
    \includegraphics[width=0.5\textwidth]{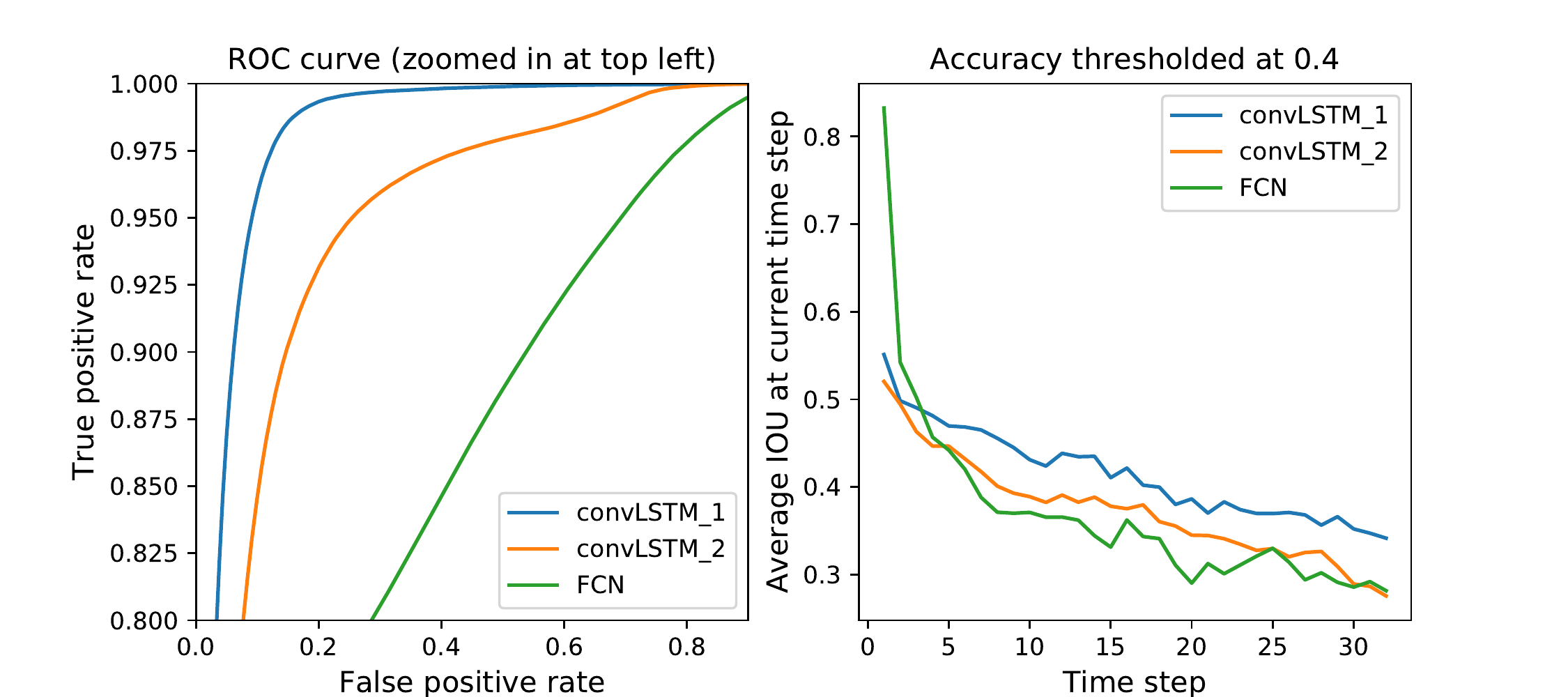}
  %  \caption{ROC comparision}

%\end{subfigure}
%\begin{subfigure}[t]{.45\textwidth}
%   \centering
%   \includegraphics[width=0.9\textwidth]{./imgs/IOU_curve.pdf}
%\caption{IOU comparison}
%   \label{fig:IOU comparison}
%\end{subfigure}
\caption{Comparison between ConvLSTM and framewise segmentation. \textbf{Left}: ConvLSTM 1 and 2 represents training strategies with and without randomly replaced frame respectively. \textbf{Right}: IOU comparison per frame.}
    \label{Local accuracy}.
\end{figure}

 We further conducted another experiment to demonstrate the benefit of using convLSTM. In this experiment, we fine tune a pre-trained segmentation network using fully convolutional neural network (FCN)\cite{long2015fully} structure for the local segmentation task. The FCN is pretrained on COCO dataset \cite{lin2014microsoft}. The feature extraction part of our local model use the same model up to pool4. When using the FCN segmentation network on a testing video, we fine tune it on the first frame of testing video clips with small learning rate and few iterations, and apply the refined model to subsequent frames. We compare the segmentation accuracy for the following 32 frames in all testing video clips. For convLSTMs trained with and without step 4, we don't fine tune based on the first frame of the testing video. We report the ROC curve and framewise IOU curve for 1200 randomly sampled video clips in the testing set in Fig. ~\ref{Local accuracy}. True positive rate and false positive rate is defined at the pixel level. Framewise IOU is defined as in Eq.~(\ref{eq.IOU}). Convolution LSTM trained under both strategies get higher AUC for the ROC curves and the FCN network with refinement during testing stage could not adapt to appearance shift as demonstrated with Fig.~\ref{Local accuracy}.

\begin{equation}
IOU(t) = \frac{1}{N_t} \sum^{N_t}_{i=1}  \frac{A_{it}^G \cap A_{it}^P}{A_{it}^G \cup A_{it}^P}
\label{eq.IOU}
\end{equation}
The better peformance using ConvLSTM for local model comes with a price, as analyzed in subsection \ref{local_framework}. ConvLSTM is not shift equivariant, a large spatial drift between consecutive frames could cause loss of tracking. In our observation, spatial shift larger than 30 pixels in the ROI could cause instability in our tracking system. To circumvent this problem, we predict the ROI using a global attention network.

%\begin{figure}
%	\begin{subfigure}[t]{0.45\textwidth}
%	        \centering
%		\includegraphics[width=0.45\textwidth]
%		\caption{area under curve}
%	\end{subfigure} %
%	\begin{subfigure}[t]{0.45\textwidth} 
%		        \centering
%		\includegraphics[width=0.45\textwidth]{./imgs/curve_compare.pdf}
%		\caption{average IOU}
%	\end{subfigure} %
%	
%
%\label{Local accuracy}
%\end{figure} 

\section{Global Attention Network} \label{Global attention network}
\begin{figure}
        \centering
	
		\includegraphics[width=0.45\textwidth]{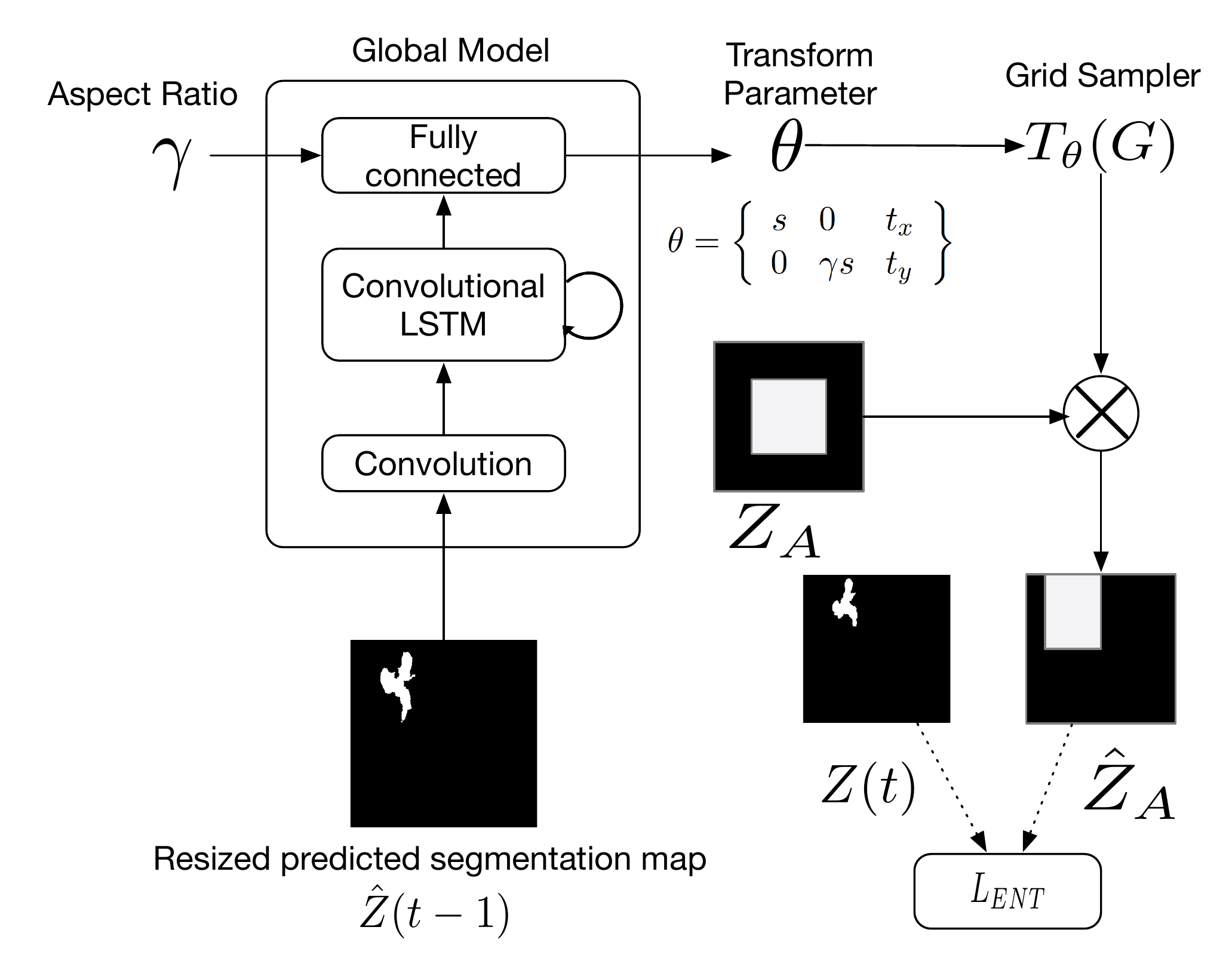}
		\caption{Training framework for global attention model}

\label{Global framework}
\end{figure}

To predict where the ROI should be located in the current frame based on predicted segmentation map in the history, one naive way is to use weighted average of the past predicted location directly to decide where the ROI should be cropped. However during the test time, the local predictor might make prediction mistakes caused by light condition, drastic appearance change, motion blur etc. Such mistakes could then cause the global model to locate a wrong ROI for the next frame. Overtimes, the ROI could drift away from the correct object location. Therefore, we need to develop a rather robust global model that can handle such problems. The ROI is specified by a spatial transform acting on a fixed anchor (a square region) $Z^A$.  We apply a LSTM on the past global segmentation maps to generate features that are then fed to a spatial transformer network to determine the transform parameter. Our spatial transform network is a special form of \cite{DBLP:journals/corr/JaderbergSZK15}, but the transformation is not applied on the feature map, but on a fixed anchor $Z_A$. The training framework of global attention network is shown in Fig.~\ref{Global framework}. 

During training stage, at each time $t$ a fixed size segmentation map $Z_{t-1}$ is feed into the global attention model $\tau$. The network generates a special form of affine transform parameters $\theta$. The spatial transform $T(\theta)$ is applied on $Z^A$, so that the transformed anchor $\hat{Z}^A_t$ maximally overlaps with the ground truth segmentation map in frame $t$, $Z_t$. We want the transformed anchor to enclose as much foreground pixel as possible, and we use a weighted $l2$ loss between $\hat{Z}^A_t$ and $Z_t$. We further add a $l2$ loss term between $\theta_t$ and $\theta_{t-1}$ so that the tranformer is temporally smooth. Parameter $\theta$ constrains the transform to only allow spatial shift and resizing. The resizing operation takes consideration of image aspect ratio, so that when cropping the image at the image domain the aspect ratio is not distorted(the ROI on the real image is always a square but with varying sizes). The overall loss function is defined as:

\begin{equation}
\begin{aligned}
&\min \limits_{\phi} \sum_{t=i}^T (L(\hat{Z}^A_{t},Z_{t}) + \lambda ||\theta_{t}-\theta_{t-1}||_2 )\\
&\hat{Z}^A_t =T(\theta_t)(Z_A)\\
&\theta_t = \tau_\phi(Z_{0:t-1})\\
\label{optimization2}
\end{aligned}
\end{equation}

The detailed number of parameters of our global model is shown in Tab.~\ref{num_parameters}. During training, we observe that recurrent model needs burn-in time to accurately predict the spatial transform. Otherwise it would not utilize the full history of the observations. So we only compute the loss after $i$th frame. In our experiment, we find setting $i=5$ works best for a total sequence length of $32$. To let our model converge faster, in practice we apply a dilation kernel on our input sequence $Z_{1:t}$ and gradually shrink the size of the dilation kernel until convergence.  

\begin{figure}
        \centering
	
		\includegraphics[width=0.45\textwidth]{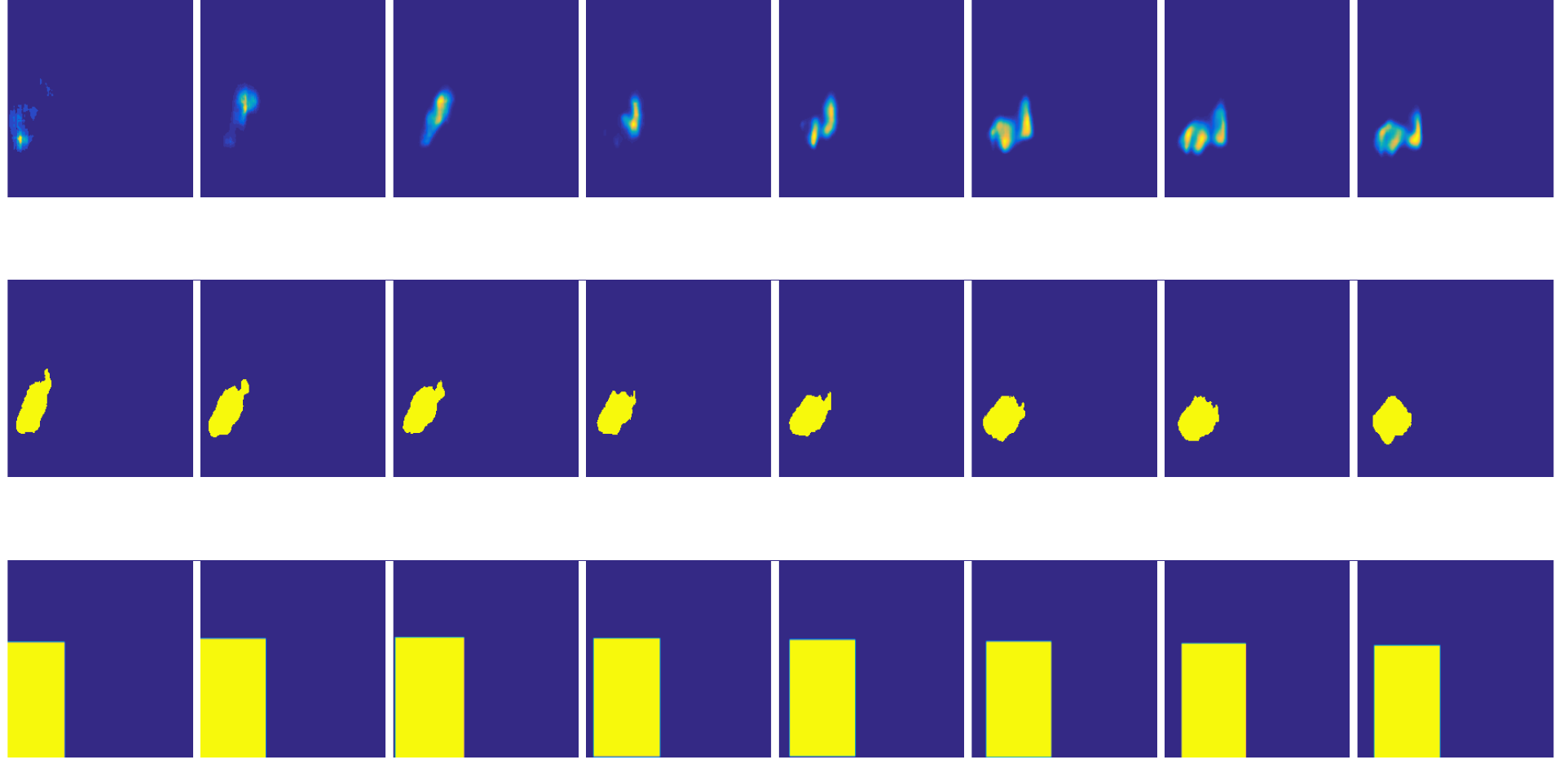}
		\caption{Demonstration of observation difference between testing set observation and ground truth. Each row shows a sequence temporally downsampled by 4. From top to bottom: input to the global model in testing sequence, ground truth mask and predicted ROI location.}

\label{training_testing_bias}
\end{figure}

 However during inference stage, since the model could only utilize the predicted masks by the local model , there is a distribution difference between testing sequences and training sequences. Fig.~\ref{training_testing_bias} demonstrates the training set and testing set difference. To handle the distribution gap we iteratively adapt our global model and local model. We discribe the way to update our model in Sec.~\ref{last_experiment}.

%
%To preserve the semetic information as much as possible,  it is always desirable to keep the aspect ratio of the object at the cropped template during testing stage. On the other hand, at training stage, the input to the global attention model is the resized version of full segmention map, which means 
% the cropped template at raw image level to preserve the aspect ratio of the object. But at the same 

%During training time, at each time a fixed sized segmenation map is feed into the 
%
%We use the history of predicted image segmentation map to predict where the template should be cropped in the current frame. One naive way is to use weighted average of the past predicted location directly to decide where the template should be cropped. However since during the test time, the local predictor might make prediction mistakes caused by light condition, drastic appearance chage, motion blur etc. Such cases could cause for template cropping location drifting away from the correct object location.

\section{Experiment} \label{last_experiment}

\subsection{Iterative Optimization}
In addition to preparing local samples as described in section~\ref{local_training}, to handle the observation difference mentioned in Sec.~\ref{Global attention network}, on each of the data fold we perform our training as following:

\begin{enumerate}
\setcounter{enumi}{0}
\item Evenly seperate video sequences of each training set into two subsets. On each subset use the ground truth bounding boxes to prepare a training set for the local model (see section~\ref{local_training}). Train one local model on each subset with early termination with loss function  Eq.~(\ref{optimization1}).

\item Train the initial global model using sequences of ground truth segmentation maps with loss function Eq.~(\ref{optimization2}). To increase convergence speed, we apply dilation operation on $Z_t$ and shrink dilation kernel size every ten thousand iterations until convergence.

\item Use the trained local model 1 from step 1 and global model from step 2 to generate predicted segmentation maps$\hat{Z}_{1:T}$, and ROI images $x_{1:T}$ and ROI segmentation maps $z_{1:T}$ for training data in subset 2. Procedure is discribed at algorithm~\ref{two_stage_algorithm}. Use local model 2 to do the same on subset 1.

\item Update the global model with modified input sequence $\bar{Z}_{0:T} :\{Z_0, \hat{Z_1},\cdots, \hat{Z_T}\}$ generated by step 3 using Eq.~(\ref{optimization2}). 

\item Train local model using the ROI image sequence $x_{1:t}$ and segmentation map sequence $z_{1:t}$, which are generated by the updated global model for the entire training set with Eq.~(\ref{optimization1}).

%use all video clips of $x_{1:t}$, $z_{1:t}$ with Eq.~\ref{local_training}

\end{enumerate}

%\begin{pseudocode}[<ovalbox>]{}{}
%\begin{algorithm}
%\caption{my algorithm}
%\begin{algorithmic}[1]
%Input:
%	First frame ground truth mask, Image: $Z_0, X_0$ 
%	Global model $M_1$, Local model $M_2$
%
%\end{algorithmic}
%\end{algorithm}

\begin{algorithm}
  \caption{Two stage tracking algorithm}
  \begin{algorithmic}[1]
    \Require{ 
Raw image $X_0$, segmentation map $Z_0$,
global model $M_1$, local model $M_2$}
\Ensure{ 
Predicted segmentation map $\hat{Z}_{1:t}$, ROI image and segmentation map sequences $x_{1:t}$, $z_{1:t}$}
    \State
Crop ROI $x_0, z_0$ with spatial paramter $\theta_0$ \\
Initialize the memory of local model $M_2$ using $x_0, z_0$\\
\textbf{for $ t = {1, T} $ do} \\
\qquad Estimate $\theta_t$ using $M_1(\hat{Z}_{t-1})$\\
\qquad  Update $\theta_{est}$ as  $\theta_{est} = \beta*\theta_t + (1-\beta)\theta_{old}$ \\
 \qquad  Get ROI images $x_t$, $z_t$ from frame $X_t$,$Z_t$ use $\theta_{est}$\\
 \qquad  Estimate $\hat{z}_t$ use $M_2(x_t)$ \\
\qquad   Use $\theta_t$ to fill in $\hat{Z}_t$ with $\hat{z}_t$

  \end{algorithmic}
\label{two_stage_algorithm}
\end{algorithm}

\begin{figure*}[h]
        \centering
	
		\includegraphics[width=1.0\textwidth]{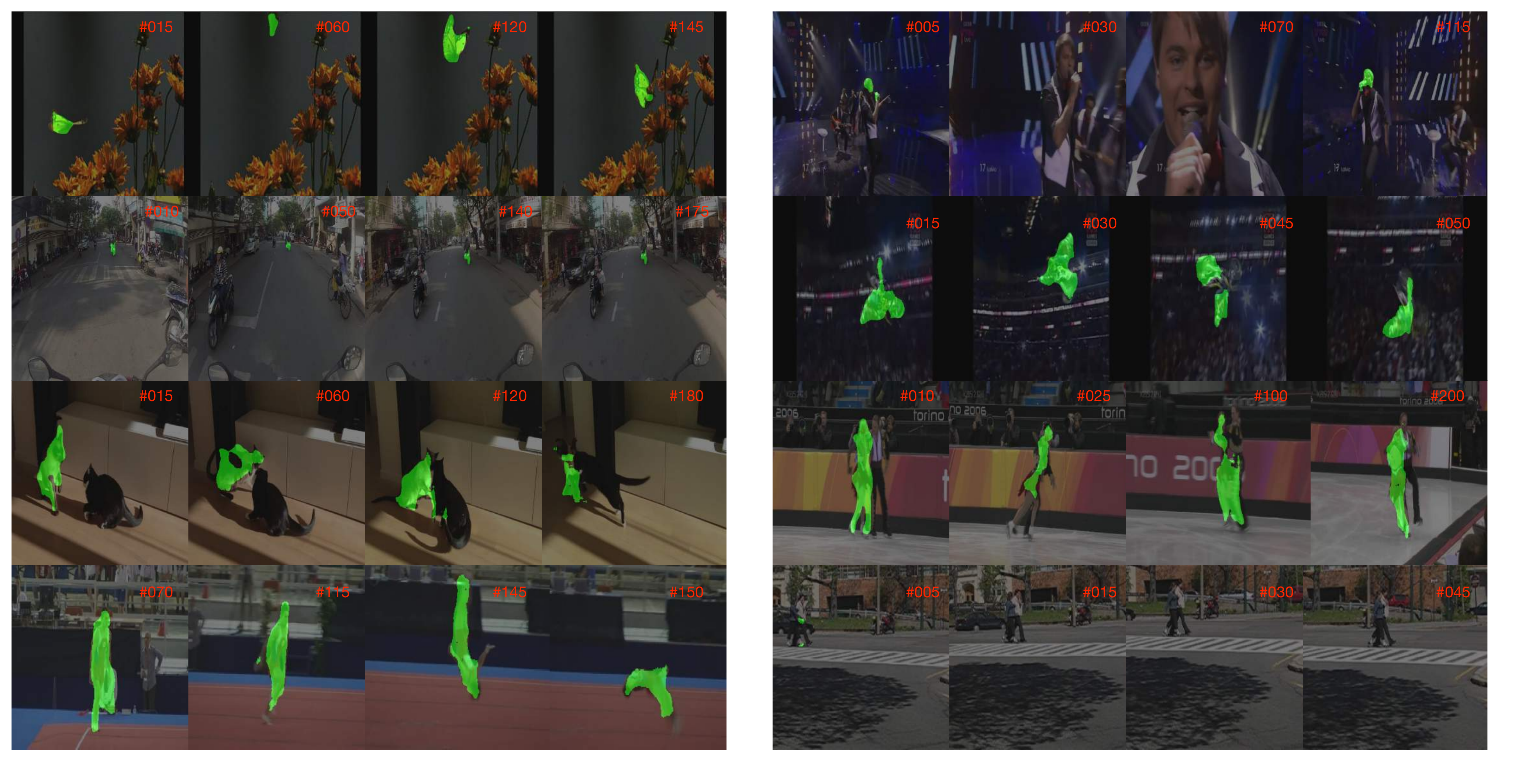}
		\caption{Tracking results for 8 videos. The predicted segmentation maps are overlaid on top of the original image. The results are obtained with the trained 2-stage model. The failure case at sequence 2 (top right) and 8 (bottom right) are mostly due to large camera motions.}

\label{Eventual_result}
\end{figure*}

\subsection{Time Complexity}
We evaluate our tracking algorith on VOT2016 \cite{unknown} segmentation dataset. We implement our algorithm with tensorflow and test it on a single NVIDIA Tesla K80 with 24G RAM. The inference speeds using the local and global models are 24 ms and 6 ms per frame respectively. With memory initialization and ROI interpolation included, the entire framework still runs more than 20 fps. 
\subsection{Quantitaive Analysis}

First, we compare our local segmentation network using LSTM with FCN segmentation network. Both models are provided with the same ROI sequences. The only difference is FCN network is further fine tuned on the first frame for each of the sequence. We follow the same procedure in Sec.  ~\ref{local_training} using the ground truth label to prepare sequences. The only exception here is that we use a exponential weighting on the ROI location center by $L_{new} = (1-\beta)L_{old}+ \beta L_{t-5}$. The decay rate $\beta$ is set at 0.8. Here, we use the object ground truth location at frame $t-5$ to crop the local ROI so that object is reasonably far apart from the center but still in the ROI. We admit that the test could be still favoring our convolutional LSTM as the location of the object is registered to be close to the center of the ROI. On the other hand this test demonstrates the upper bound of the proposed object tracking approach, achievable when the global model can accurately locate the ROI. The evaluation using the FCN segmentation network, on the other hand, is meant to evaluate the achievable tracking performance by a CNN-based segmentation network, when equipped with a near perfect global tracker. The result is shown in Tab.~\ref{result1}.

Next, we evaluate our global model and local model jointly. The inference for the two stage model (denoted by 2-stage ConvLSTM) follows algorithm~\ref{two_stage_algorithm}. We also evaluate a benchmark 1-stage model, which replaces the global model with a simple predictor for the ROI center,  described in Eq.~(\ref{weighted_location}). Here $p_i$ is the estimated probalility of pixel $i$ belonging to the foreground by the local model for the previous frame. The ROI size is fixed. FCN based tracking uses the same approach for determining the ROI with its own segmentation map. Table~\ref{result2} compares the performance of our 2-stage model, the benchmark 1-stage and the FCN segmentation network.

%demonstrate the result of using both models and single model. We further provide another CNN based segmentation benchmark in table.~\ref{result2}, the estimated location is using the predicted probability of the segmentation network.

Results in Tables \ref{result1} and~\ref{result2} demonstrate that the proposed 2-stage convLSTM architecture is better than a CNN fine tuned on the first frame. Even when provided with nearly correct location of the ROI in each new frame, the CNN-based segmentation network could not handle appearance shift as well as our 2-stage ConvLSTM. Furthermore, our global model performs better than a naive location predictor. Sample visual results are shown in Fig.~\ref{Eventual_result}.

\begin{equation}
\begin{aligned}
\theta_x =\frac{1}{\sum_i^I p_i}\sum_i^I p_i x_i \quad \theta_y =\frac{1}{\sum_i^I p_i}\sum_i^I p_i y_i 
\end{aligned}
\label{weighted_location}
\end{equation}
 
\begin{table}
\scalebox{0.8}{
\begin{tabular}{  p{3cm} | p{1.2cm} p{1.2cm} p{1.2cm} p{1.2cm} }
\toprule       
Test sequence length threshold at 0.4 &20  &40 & 80 & 160 \\
\toprule     
 ConLSTM & 0.4690 &   0.4405   & 0.3962  &  0.3342 \\
FCN & 0.3785  & 0.3582&  0.3157 & 0.2304\\

\bottomrule

 \end{tabular}}
\caption{Local segmentation network evaluation: average IOU at different sequence length using ROIs that are close to ground truth location. For sequences shorter than the preset length, we upsample the testing sequences to the fixed length.}
\label{result1}
 \end{table}

\begin{table}
\scalebox{0.8}{
\begin{tabular}{  p{3cm} | p{1.2cm} p{1.2cm} p{1.2cm} p{1.2cm} }
\toprule       
Test sequence length threshold set at 0.4 &20  &40 & 80 & 160 \\
\toprule     
2-stage ConLSTM  & 0.3992&  0.3606   &  0.3201    &  0.2564\\
1-stage ConLSTM & 0.380& 0.34601 & 0.3046  &  0.2419 \\
FCN & 0.2275   &   0.2058 &  0.1678 & 0.1437 \\
\toprule       
Test sequence length threshold  at 0.7 &20  &40 & 80 & 160 \\
\toprule
2-stage ConLSTM  &  0.2485  &     0.2302 &  0.202 &  0.1854 \\
1-stage ConLSTM  & 0.2080 &  0.1926 &  0.1679 & 0.1518 \\ 
 FCN & 0.1030   &0.093 & 0.0711 & 0.058 \\
\bottomrule

 \end{tabular}}
\caption{Overall network evaluation: average IOU at different sequence length.}
\label{result2}
 \end{table}

\section{ACKNOWLEDGEMENT}
This work was funded by National Science Foundation award CCF-1422914.
 \vspace{-0.05in}

\section{Conclusion}
In this work, we tackle tracking problem at the pixel level. By providing the beginning frame and corresponding segmentation map, we model the appearance shift as a time series. We propose a novel two-stage model handling micro-scale appearance change and macro-scale object motion seperately. The local segmentation model has far better performance compared to a CNN fine-tuned on the first frame. The global model can accurately predict the rough location and size of the object from frame to frame. We demonstrate our novel approach on a very challenging VOT dataset. Finally our model performs pixel-wise object tracking at a reasonable accuracy in real time.

{\small
\bibliographystyle{ieee}
\bibliography{mybib}

\begin{thebibliography}{10}\itemsep=-1pt

\bibitem{aeschliman2010probabilistic}
C.~Aeschliman, J.~Park, and A.~C. Kak.
\newblock A probabilistic framework for joint segmentation and tracking.
\newblock In {\em Computer Vision and Pattern Recognition (CVPR), 2010 IEEE
  Conference on}, pages 1371--1378. IEEE, 2010.

\bibitem{belagiannis2012segmentation}
V.~Belagiannis, F.~Schubert, N.~Navab, and S.~Ilic.
\newblock Segmentation based particle filtering for real-time 2d object
  tracking.
\newblock {\em Computer Vision--ECCV 2012}, pages 842--855, 2012.

\bibitem{bertinetto2016fully}
L.~Bertinetto, J.~Valmadre, J.~F. Henriques, A.~Vedaldi, and P.~H. Torr.
\newblock Fully-convolutional siamese networks for object tracking.
\newblock In {\em European Conference on Computer Vision}, pages 850--865.
  Springer, 2016.

\bibitem{everingham2015pascal}
M.~Everingham, S.~A. Eslami, L.~Van~Gool, C.~K. Williams, J.~Winn, and
  A.~Zisserman.
\newblock The pascal visual object classes challenge: A retrospective.
\newblock {\em International journal of computer vision}, 111(1):98--136, 2015.

\bibitem{gan2015first}
Q.~Gan, Q.~Guo, Z.~Zhang, and K.~Cho.
\newblock First step toward model-free, anonymous object tracking with
  recurrent neural networks.
\newblock {\em arXiv preprint arXiv:1511.06425}, 2015.

\bibitem{girshick2015fast}
R.~Girshick.
\newblock Fast r-cnn.
\newblock In {\em Proceedings of the IEEE international conference on computer
  vision}, pages 1440--1448, 2015.

\bibitem{pmlr-v37-hong15}
S.~Hong, T.~You, S.~Kwak, and B.~Han.
\newblock Online tracking by learning discriminative saliency map with
  convolutional neural network.
\newblock In F.~Bach and D.~Blei, editors, {\em Proceedings of the 32nd
  International Conference on Machine Learning}, volume~37 of {\em Proceedings
  of Machine Learning Research}, pages 597--606, Lille, France, 07--09 Jul
  2015. PMLR.

\bibitem{DBLP:journals/corr/JaderbergSZK15}
M.~Jaderberg, K.~Simonyan, A.~Zisserman, and K.~Kavukcuoglu.
\newblock Spatial transformer networks.
\newblock {\em CoRR}, abs/1506.02025, 2015.

\bibitem{jang2017online}
W.-D. Jang and C.-S. Kim.
\newblock Online video object segmentation via convolutional trident network.
\newblock In {\em Proceedings of the IEEE Conference on Computer Vision and
  Pattern Recognition}, pages 5849--5858, 2017.

\bibitem{kahou2015ratm}
S.~E. Kahou, V.~Michalski, and R.~Memisevic.
\newblock Ratm: recurrent attentive tracking model.
\newblock {\em arXiv preprint arXiv:1510.08660}, 2015.

\bibitem{kingma2013auto}
D.~P. Kingma and M.~Welling.
\newblock Auto-encoding variational bayes.
\newblock {\em arXiv preprint arXiv:1312.6114}, 2013.

\bibitem{unknown}
M.~Kristan, A.~Leonardis, J.~Matas, M.~Felsberg, R.~Pflugfelder, L.~Čehovin,
  T.~Vojír, G.~Häger, A.~Lukežič, G.~Fernandez~Dominguez, A.~Gupta,
  A.~Petrosino, A.~Memarmoghadam, A.~Garcia-Martin, A.~Solís~Montero,
  A.~Vedaldi, A.~Robinson, A.~Ma, A.~Varfolomieiev, and Z.~Chi.
\newblock The visual object tracking vot2016 challenge results, 10 2016.

\bibitem{lin2014microsoft}
T.-Y. Lin, M.~Maire, S.~Belongie, J.~Hays, P.~Perona, D.~Ramanan,
  P.~Doll{\'a}r, and C.~L. Zitnick.
\newblock Microsoft coco: Common objects in context.
\newblock In {\em European conference on computer vision}, pages 740--755.
  Springer, 2014.

\bibitem{long2015fully}
J.~Long, E.~Shelhamer, and T.~Darrell.
\newblock Fully convolutional networks for semantic segmentation.
\newblock In {\em Proceedings of the IEEE Conference on Computer Vision and
  Pattern Recognition}, pages 3431--3440, 2015.

\bibitem{makhzani2015adversarial}
A.~Makhzani, J.~Shlens, N.~Jaitly, I.~Goodfellow, and B.~Frey.
\newblock Adversarial autoencoders.
\newblock {\em arXiv preprint arXiv:1511.05644}, 2015.

\bibitem{nam2016modeling}
H.~Nam, M.~Baek, and B.~Han.
\newblock Modeling and propagating cnns in a tree structure for visual
  tracking.
\newblock {\em arXiv preprint arXiv:1608.07242}, 2016.

\bibitem{nam2016learning}
H.~Nam and B.~Han.
\newblock Learning multi-domain convolutional neural networks for visual
  tracking.
\newblock In {\em Proceedings of the IEEE Conference on Computer Vision and
  Pattern Recognition}, pages 4293--4302, 2016.

\bibitem{ning2017spatially}
G.~Ning, Z.~Zhang, C.~Huang, X.~Ren, H.~Wang, C.~Cai, and Z.~He.
\newblock Spatially supervised recurrent convolutional neural networks for
  visual object tracking.
\newblock In {\em Circuits and Systems (ISCAS), 2017 IEEE International
  Symposium on}, pages 1--4. IEEE, 2017.

\bibitem{perazzi2016benchmark}
F.~Perazzi, J.~Pont-Tuset, B.~McWilliams, L.~Van~Gool, M.~Gross, and
  A.~Sorkine-Hornung.
\newblock A benchmark dataset and evaluation methodology for video object
  segmentation.
\newblock In {\em Proceedings of the IEEE Conference on Computer Vision and
  Pattern Recognition}, pages 724--732, 2016.

\bibitem{pinheiro2015learning}
P.~O. Pinheiro, R.~Collobert, and P.~Doll{\'a}r.
\newblock Learning to segment object candidates.
\newblock In {\em Advances in Neural Information Processing Systems}, pages
  1990--1998, 2015.

\bibitem{redmon2016you}
J.~Redmon, S.~Divvala, R.~Girshick, and A.~Farhadi.
\newblock You only look once: Unified, real-time object detection.
\newblock In {\em Proceedings of the IEEE Conference on Computer Vision and
  Pattern Recognition}, pages 779--788, 2016.

\bibitem{ren2015faster}
S.~Ren, K.~He, R.~Girshick, and J.~Sun.
\newblock Faster r-cnn: Towards real-time object detection with region proposal
  networks.
\newblock In {\em Advances in neural information processing systems}, pages
  91--99, 2015.

\bibitem{romera2016recurrent}
B.~Romera-Paredes and P.~H.~S. Torr.
\newblock Recurrent instance segmentation.
\newblock In {\em European Conference on Computer Vision}, pages 312--329.
  Springer, 2016.

\bibitem{simonyan2014very}
K.~Simonyan and A.~Zisserman.
\newblock Very deep convolutional networks for large-scale image recognition.
\newblock {\em arXiv preprint arXiv:1409.1556}, 2014.

\bibitem{son2015tracking}
J.~Son, I.~Jung, K.~Park, and B.~Han.
\newblock Tracking-by-segmentation with online gradient boosting decision tree.
\newblock In {\em Proceedings of the IEEE International Conference on Computer
  Vision}, pages 3056--3064, 2015.

\bibitem{tao2016siamese}
R.~Tao, E.~Gavves, and A.~W. Smeulders.
\newblock Siamese instance search for tracking.
\newblock In {\em Proceedings of the IEEE Conference on Computer Vision and
  Pattern Recognition}, pages 1420--1429, 2016.

\bibitem{wang2016stct}
L.~Wang, W.~Ouyang, X.~Wang, and H.~Lu.
\newblock Stct: Sequentially training convolutional networks for visual
  tracking.
\newblock In {\em Proceedings of the IEEE Conference on Computer Vision and
  Pattern Recognition}, pages 1373--1381, 2016.

\bibitem{wang2015transferring}
N.~Wang, S.~Li, A.~Gupta, and D.-Y. Yeung.
\newblock Transferring rich feature hierarchies for robust visual tracking.
\newblock {\em arXiv preprint arXiv:1501.04587}, 2015.

\bibitem{xingjian2015convolutional}
S.~Xingjian, Z.~Chen, H.~Wang, D.-Y. Yeung, W.-k. Wong, and W.-c. Woo.
\newblock Convolutional lstm network: A machine learning approach for
  precipitation nowcasting.
\newblock In {\em Advances in Neural Information Processing Systems}, pages
  802--810, 2015.

\bibitem{yang2016object}
J.~Yang, B.~Price, S.~Cohen, H.~Lee, and M.-H. Yang.
\newblock Object contour detection with a fully convolutional encoder-decoder
  network.
\newblock In {\em Proceedings of the IEEE Conference on Computer Vision and
  Pattern Recognition}, pages 193--202, 2016.

\bibitem{yang2017recurrent}
T.~Yang and A.~B. Chan.
\newblock Recurrent filter learning for visual tracking.
\newblock {\em arXiv preprint arXiv:1708.03874}, 2017.

\bibitem{yeo2017superpixel}
D.~Yeo, J.~Son, B.~Han, and J.~Hee~Han.
\newblock Superpixel-based tracking-by-segmentation using markov chains.
\newblock In {\em Proceedings of the IEEE Conference on Computer Vision and
  Pattern Recognition}, pages 1812--1821, 2017.

\bibitem{DBLP:journals/corr/ZeilerF13}
M.~D. Zeiler and R.~Fergus.
\newblock Visualizing and understanding convolutional networks.
\newblock {\em CoRR}, abs/1311.2901, 2013.

\end{thebibliography}
}

\end{document}